\documentclass[journal]{journal}

\usepackage{subfigure}
\usepackage{url}
\usepackage{amsthm}
\usepackage{amsmath}
\usepackage{amssymb}

\usepackage[sort]{cite}
\usepackage{hyperref}

\usepackage{graphicx}
\usepackage{microtype}
\usepackage{graphicx}
\usepackage{subfigure}
\usepackage{booktabs} % for professional tables
\usepackage{oplotsymbl}

\usepackage{tikz}

\usepackage{algpseudocode}
\usepackage{algorithm}
\usepackage{algorithm}
\usepackage{algpseudocode}
\usepackage{eqparbox}
\newdimen{\algindent}
\algnewcommand\algorithmicforeach{\textbf{for each}}
\algdef{S}[FOR]{ForEach}[1]{\algorithmicforeach\ #1\ \algorithmicdo}
\algnewcommand\LeftCommentNoIntent[1]{%
$\triangleright$ \eqparbox{COMMENT}{#1} \hfill %
}

\newcommand{\normtwo}[1]{\left\lVert#1\right\rVert_{2}}

\newcommand{\norm}[1]{\left\lVert#1\right\rVert_\textrm{2}}
\newtheorem{innercustomgeneric}{\customgenericname}
\providecommand{\customgenericname}{}
\newcommand{\newcustomtheorem}[2]{%
  \newenvironment{#1}[1]
  {%
   \renewcommand\customgenericname{#2}%
   \renewcommand\theinnercustomgeneric{##1}%
   \innercustomgeneric
  }
  {\endinnercustomgeneric}
}

\newcustomtheorem{customthm}{Theorem}
\newcustomtheorem{customlemma}{Lemma}

\begin{document}
%\title{Bare Demo of IEEEtran.cls for Journals}
\title{On a Conjecture Regarding\\ the Adam Optimizer\vspace{1cm}}
%\iffalse
\author{Mohamed~Akrout, Douglas Tweed% <-this % stops a space
\thanks{M. Akrout is with AIP Labs, Budapest, Hungary.}
\thanks{D. Tweed is with the Department of Physiology,
University of Toronto, Toronto, Canada.}
\thanks{Correspondence to: Douglas Tweed $<$douglas.tweed@utoronto.ca$>$}
}

\maketitle
\thispagestyle{empty}

\begin{abstract}
%\boldmath
Why does the Adam optimizer work so well in deep-learning applications? Adam's originators, Kingma and Ba, presented a mathematical argument that was meant to help explain its success, but Bock and colleagues have since reported that a key piece is missing from that argument — an unproven lemma which we will call Bock’s conjecture. Here we show that this conjecture is false, but we prove a modified version of it — a generalization of a result of Reddi and colleagues — which can take its place in analyses of Adam.
\end{abstract}


% \fi
\IEEEpeerreviewmaketitle

\section{Introduction}

Kingma and Ba \cite{kingma2014adam} tried to prove that their Adam optimizer zeroed the error-measure known as \textit{average regret}, in a learning task called \textit{online convex optimization} \cite{zinkevich2003online}. Rubio \cite{rubio2017convergence} and Bock et al. \cite{bock2018improvement} found mistakes in the proof, and Bock et al. managed to repair most of them, but they could not verify one key statement, called Lemma 10.4 in Kingma and Ba's paper and Conjecture 4.2 in Bock's. We will show that this conjecture is in fact false, but that a modified version of it does hold. This modified version generalizes an earlier result proven by Reddi and colleagues for their AMSGrad optimizer \cite{reddi2019convergence}, so our result can replace Bock’s Conjecture in analyses of most common variants of Adam.

For tractability, analyses of Adam typically use versions of the algorithm that are slightly different from the one generally employed in deep learning. Here, we will use the version laid out in Algorithm \ref{algo:adam-optimizer}, which differs from that of Kingma, Ba, and Bock et al. only in that they set $\lambda_m = \lambda_g \in (0,1)$. We will explain the significance of this difference where it becomes relevant.

\begin{algorithm}
\caption{Adam optimizer}\label{algo:adam-optimizer}
\begin{algorithmic}[1] \vspace{0.3cm}
\Require $\eta > 0; \beta_1, \beta_2 \in (0,1)$; $\lambda_m, \lambda_g \in (0,1]$; duration $T \in \mathbb{Z}^+$; initial parameter (weight and bias) vector $\theta_{0}$; convex differentiable loss functions $f_t(\theta)$; $m_0, v_0 = 0$.\vspace{0.1cm}
\Statex\hspace{-0.6cm}\textbf{Return:} updated parameter vector $\theta_{T}$.\vspace{0.2cm}
\State \textbf{for} $t = 1$ to $T$ \textbf{do}
\State \hspace{0.3cm}$g_t = \nabla_\theta\,f_t(\theta_{t-1})$\label{algo:gt}\vspace{0.1cm}
\Statex\hspace{0.3cm}\LeftCommentNoIntent{Compute biased moment estimates}
\State \hspace{0.3cm}$m_t = \beta_{1}\,\lambda_{m}^{t-1}\,m_{t-1} + (1-\beta_{1}\lambda_{g}^{t-1})\,g_{t}$\label{algo:mt}
\State \hspace{0.3cm}$v_t = \beta_{2}\,v_{t-1} + (1-\beta_{2})\,g_{t}^2$\label{algo:vt}\vspace{0.1cm}
\Statex\hspace{0.3cm}\LeftCommentNoIntent{Bias-correct the moment estimates}
\State \hspace{0.3cm}$\widehat{m}_t = {m_t}/{(1-\beta_1^t)}$\label{algo:mt-hat}
\State \hspace{0.3cm}$\widehat{v}_t = {v_t}/{(1-\beta_2^t)}$\label{algo:vt-hat}\vspace{0.1cm}
\Statex\hspace{0.3cm}\LeftCommentNoIntent{Update the parameters}
\State \hspace{0.3cm}$\theta_{t} = \theta_{t-1} - (\eta/\sqrt{t})~ \widehat{m}_t/\sqrt{\widehat{v}_t}$\label{algo:thetat}
\State \textbf{end}
\newline
\end{algorithmic}
\end{algorithm}

In this algorithm, each of the variables $g_t$, $m_t$, $v_t$, $\widehat{m}_t$, $\widehat{v}_t$, and $\theta_t$ is a real-valued vector; for instance $g_{t}$ is the $g$-vector at time $t$. But the vector operations in Adam are all element-wise, except possibly in line \ref{algo:gt}, and therefore we can analyse the parts after that line element-wise, i.e. we can assume throughout this paper that the vectors $g_t$ etc. have one element each (except in Section 4, where calculations of regret depend on non-element-wise operations outside Adam). We will write $g_{1:T}$ for the $T$-element vector $[g_1, g_2, ..., g_T]$. We also define
\begin{equation}\label{eq:x-definitions}
    x_1 \triangleq 1 - \beta_1,~ x_2 \triangleq 1 - \beta_2,
\end{equation}
and
\begin{equation}\label{eq:s-definition}
    s_T \triangleq \sum_{t=1}^{T} \frac{\widehat{m}_t^2}{\sqrt{t \,\widehat{v}_t}},
\end{equation}
which is central to Bock's conjecture. We assume $g_{1} \neq 0$ because otherwise $s_{T}$ is undefined. 

\noindent We can now state\\

\textbf{Bock's conjecture.} \textit{In Algorithm} \ref{algo:adam-optimizer}\textit{, if $\lambda_m = \lambda_g \in (0,1)$ and $\gamma = \beta_1^2/\sqrt{\beta_2} < 1$ then for any $g_{1:T}$~we have}
\begin{equation}\label{eq:bock-conjecture}
    s_T \leq \frac{2}{(1-\gamma)} \frac{1}{\sqrt{1-\beta_{2}}}\normtwo{g_{1: T}}.
\end{equation}

We will call the right-hand side of this inequality the \textit{Kingma-Ba} or \textit{K-B bound}.

\section{Counterexample to Bock's conjecture}\label{sec:counterexample}

Consider vectors $g_{1:T}$ where $g_t > 0 ~\forall t$. Set $\beta_{1}$ and $\beta_{2}$ equal, i.e. $\beta_{1} = \beta_{2} = \beta$, and observe that $s_T$ and the K-B bound are right-continuous functions of $\beta$ at $\beta = 0$. So if we can find a counterexample where $\beta = 0$ then there also exist counterexamples where $\beta \in (0,1)$.

Letting $\beta \rightarrow 0$, we get $\widehat{m}_t = g_t$ and $\widehat{v}_t = g_t^2$ (from lines \ref{algo:mt}--\ref{algo:vt-hat} of Algorithm~\ref{algo:adam-optimizer}) and $\gamma = \beta^{3/2} = 0$. Bock's conjecture then takes the form

\begin{equation*}
    \sum_{t=1}^T \frac{g_t}{\sqrt{t}} \leq 2\normtwo{g_{1: T}}.
\end{equation*}

If we choose $g_{t} = 1/\sqrt{t}$ then this inequality becomes

\begin{equation*}
    \sum_{t=1}^{T} \frac{1}{t} \leq 2\sqrt{\sum_{t=1}^{T} \frac{1}{t}},
\end{equation*}

\noindent which is false when the left-hand side $> 4$, as happens when $T > 30$.

\noindent By continuity, (\ref{eq:bock-conjecture}) is also violated in cases where $\beta \in (0, 1)$. This plot shows an example:
\begin{figure}[h!]
\centering
\includegraphics[scale = 0.63]{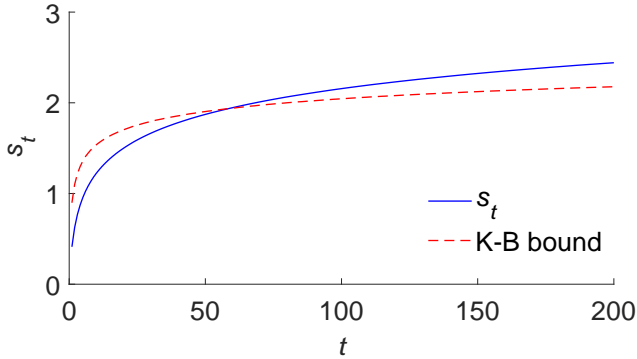}
\caption{We ran Algorithm~\ref{algo:adam-optimizer} for 200 time steps using $\beta_1 = \beta_2 = 0.1$, $\lambda_m = \lambda_g = 1 - 10^{-8}$, and $g_t = 1/\sqrt{t}$, and we computed the K-B bound and $s_t$ at each step. $s_t$ surpassed the bound at $ t = 59$.}
\label{fig:beta-0-1}
\end{figure}

\section{Modifying Bock's conjecture}
We want to replace the K-B bound on the right-hand side of (\ref{eq:bock-conjecture}) with a different bound that we can verify, at least for values of $\beta_{1}$ and $\beta_{2}$ that are typically used in AI applications of Adam.

%\newpage
%\begin{customlemma}{3.1}\label{lemma:lemma3.1}
\textbf{Lemma 1.} % using customlemma makes the typesetting weird
\textit{In Algorithm} \ref{algo:adam-optimizer}\textit{, if $\lambda_m = \lambda_g = 1$, $\rho = \beta_{2} / \beta_{1}^{2} \in (1,2)$, and $K = \rho / (\rho - 1)$ then $\forall t \in [1,\infty)$}
\begin{equation}\label{eq:lemma3.1_inequality}
    \frac{m^2_t}{v_t} < K\,\frac{x_1^2}{x_2}.
\end{equation}
%\end{customlemma}

\begin{proof}
By induction:\\\\
$(i)$ At $t=1$, we have
\begin{equation*}
    \frac{m_1^2}{v_1} = \frac{x_1^2\,g_1^2}{x_2\,g_1^2} = \frac{x_1^2}{x_2},
\end{equation*}
and so
\begin{equation*}
    \frac{m_1^2}{v_1} < K\,\frac{x_1^2}{x_2}
\end{equation*}
because $K > 1$.\\\\
$(ii)$ Next we show that if (\ref{eq:lemma3.1_inequality}) holds at any time $t$ then it still holds at $t + 1$, i.e.:
\begin{equation}\label{eq:lemma3.1_induction}
    \frac{m^2_{t+1}}{v_{t+1}}- K\,\frac{x_1^2}{x_2} < 0,
\end{equation}
or equivalently,
\begin{equation*}
    m^2_{t+1} - K\,\frac{x_1^2}{x_2}v_{t+1}\ < 0.
\end{equation*}
If we substitute the formulas for $m_{t+1}$ and $v_{t+1}$ from lines \ref{algo:mt} and \ref{algo:vt} of Algorithm~\ref{algo:adam-optimizer}, and use the definitions in (\ref{eq:x-definitions}), the left-hand side becomes
\begin{equation}\label{eq:lemma3.1-intermediate}
    (\beta_1\,m_t + x_1g_{t+1})^2 - K \,\frac{x_1^2}{x_2}\,(\beta_2\,v_t + x_2g_{t+1}^2).
\end{equation}
We expand the squared sum, rearrange, and apply (\ref{eq:lemma3.1_inequality}) to see that (\ref{eq:lemma3.1-intermediate}) is less than
\begin{equation*}
    \beta_1^2\,m_t^2 + 2\,\beta_1\,m_t\,x_1\,g_{t+1} - (K-1)\,x_1^2\,g_{t+1}^2 - \beta_2\,m_t^2.
\end{equation*}

We break up the first addend into a sum of two terms to get
\begin{align*}
    \frac{K}{K-1}\,\beta_1^2\,m_t^2 -\frac{1}{K-1}\,\beta_1^2\,m_t^2 + 2\,\beta_1\,m_t\,x_1\,g_{t+1} \\
    -(K-1)\,x_1^2\,g_{t+1}^2 - \beta_2\,m_t^2,
\end{align*}
which is
\begin{align*}
    &\left(\frac{K}{K-1}\,\beta_1^2 - \beta_2\right) m_t^2 & \\
    - &\left(\frac{1}{\sqrt{K-1}}\,\beta_1\,m_t - \sqrt{K-1}\,x_1\,g_{t+1}\right)^2.
\end{align*}

By the definition of $K$, the top line here equals $0$, and the quantity as a whole $\leq 0$, proving (\ref{eq:lemma3.1_induction}).
\end{proof}
\vspace{0.5cm}

%\begin{customlemma}{1}\label{lemma:lemma1}
\textbf{Lemma 2.}  % using customlemma makes the typesetting weird
\textit{In Algorithm} \ref{algo:adam-optimizer}\textit{, if $\lambda_m = \lambda_g = 1$ and $\beta_2 \geq 2\,\beta_1-\beta_1^2$ then $\forall t \in [1,\infty)$}
\begin{equation*}
    \frac{\widehat{m}_t^2}{\widehat{v}_t} \leq  \frac{m^2_t}{v_t}.
\end{equation*}
%\end{customlemma}

\begin{proof}
From lines \ref{algo:mt-hat} and \ref{algo:vt-hat} of Algorithm~\ref{algo:adam-optimizer} we have
\begin{equation*}\label{eq:c-definition}
    \frac{\widehat{m}_t^2}{\widehat{v}_t} = c_t\, \frac{m_t^2}{v_t}, ~~\textrm{where}~~c_t \triangleq \frac{1-\beta_2^t}{(1-\beta_1^t)^2}.
\end{equation*}
Note that $c_1 = x_2/x_1^2$, which is $\leq 1$ when $\beta_2 \geq 2\,\beta_1-\beta_1^2$. To prove that $c_t \leq 1~\forall t \in [1,\infty)$, we define this function of continuous time:
\begin{equation}\label{eq:h-definition}
    h(t) \triangleq (1-\beta_2^t) - (1-\beta_1^t)^2.
\end{equation}
We will show that $\forall t \in [1, \infty)$
\begin{equation*}
    \frac{dh}{dt} \leq 0 ~\textrm{whenever}~h(t) = 0,
\end{equation*}
because that means $h(t)$, starting at $h(1) = x_2 - x_1^2 \leq 0$, can never cross over to any positive value, and therefore $c_t$ stays $\leq 1$.

We have
\begin{align*}\label{eq:time-derivative-1}
    \frac{dh}{dt} = - \log \beta_2 + (1-\beta_2^t)\,\log \beta_2 
    + 2 \,\beta_1^t\,(1-\beta_1^t)\,\log \beta_1,
\end{align*}
and if $h(t) = 0$,
\begin{align*}
    \frac{dh}{dt} - \log \beta_2 + (1-\beta_1^t)^2\,\log \beta_2
    + 2 \,\beta_1^t\,(1-\beta_1^t)\,\log \beta_1,
\end{align*}
because then $(1-\beta_2^t) = (1-\beta_1^t)^2$ by the definition in (\ref{eq:h-definition}).

We define $\alpha \triangleq 1-\beta_1^t$ to get, $\forall \alpha\in [x_1, 1)$,
\begin{equation}\label{eq:time-derivative-2}
\begin{aligned}[b]
    \frac{dh}{dt} &= -\log\beta_2 + \log\beta_2\,\alpha^2 + 2\,\log \beta_1\,\alpha\,(1-\alpha)\\%, ~~\forall \alpha\in [x_1, 1)\\
    &=\log\beta_2\,\underbrace{\Big((1-r)\,\alpha^2 + r\,\alpha - 1\Big)}_{P(\alpha)},
\end{aligned}
\end{equation}
where $r \triangleq 2\,\log\beta_1/\log \beta_2$, which $> 1$ when $\beta_2 \geq 2\,\beta_1-\beta_1^2$. Because $r > 1$, the polynomial $P(\alpha)$ is concave down. It follows that $P(\alpha) \geq 0$ on $[x_1, 1)$, because $P(1) = 0$ and $P(x_1) \geq 0$ by the conditions on $\beta_1$ and $\beta_2$ (see the Appendix). Therefore by (\ref{eq:time-derivative-2}), $dh/dt \leq 0$ at any $t \in [1, \infty)$ where $h(t) = 0$, which means $h$ can never cross $0$ and $c_t$ can never exceed $1$.
\end{proof}

\vspace{0.3cm}
%\begin{result}\label{result:result3}
\textbf{Result 1. }
\textit{In Algorithm} \ref{algo:adam-optimizer}\textit{, if $\lambda_g = 1$, $\beta_{2} < 2\beta_{1}^{2}$, $\beta_2 \geq 2\,\beta_1-\beta_1^2$, $K = \beta_2/(\beta_2 - \beta_1^2)$ as in Lemma 3.1, and $\tau = \lfloor -log(2)/log(\beta_1) \rfloor$ then $\forall T \in [1,\infty)$}
\begin{equation}\label{eq:s-bound}
    s_T < (2+\sqrt{\tau})\,\sqrt{ 1 + K\,\frac{x_1^2}{x_2}\,\log T}~~ \normtwo{g_{1:T}}.
\end{equation}
%\end{result}
\begin{proof}
 We may assume that $\normtwo{g_{1:T}} = 1$, as $s_T$ is a homogeneous function of degree $1$ of $g_{1:T}$; that is, if we multiply every element of $g_{1:T}$ by a constant, $\zeta$, then the effect on $s_T$ is to multiply it by $\zeta$ as well. We can also say that $g_t \geq 0 ~\forall t \in [1, \infty)$, as we are seeking an \textit{upper} bound for $s_T$, and given any $g_{1:T}$ with negative elements, we could always increase $s_T$ by flipping the signs of those negative $g_t$. And we can assume that $\lambda_m = 1$, because any $\lambda_m \in (0,1)$ would only shrink $s_T$, as is clear from (\ref{eq:s-definition}), lines \ref{algo:mt} and \ref{algo:mt-hat} of Algorithm~\ref{algo:adam-optimizer}, and the non-negativity of all the $g_t$.

The definition of $s_T$ in (\ref{eq:s-definition}) shows that it is the dot product of two vectors:
\begin{equation}\label{eq:dot-product}
    s_T = \widehat{m}_{1:T} \cdot \mu_{1:T},   
\end{equation}
where $\mu_{1:T}$ is the vector with elements $\mu_t = \widehat{m}_t/\sqrt{t\widehat{v}_t}$.

The first vector in this dot product has a bounded 2-norm. First of all,
\begin{equation*}
    \normtwo{m_{1:T}} \leq \normtwo{g_{1:T}} = 1    
\end{equation*}
because $m_{1:T}$ is an exponential moving average of $g_{1:T}$, and the 2-norm of such an average cannot exceed the 2-norm of its input. Then we get each $\widehat{m}_t$ by multiplying $m_t$ by the factor $1/(1-\beta_1^t)$. For all $t > \tau$, those factors are $<2$, so $\normtwo{\widehat{m}_{\tau + 1 : T}} < 2\normtwo{m_{\tau + 1 : T}} < 2$. And $\normtwo{\widehat{m}_{1:\tau}} \leq \sqrt{\tau}$ because $\widehat{m}_t \leq 1 ~\forall t \in [1, \infty)$. Therefore
\begin{equation}\label{eq:dot-1}
    \normtwo{\widehat{m}_{1:T}} < 2 + \sqrt{\tau}.    
\end{equation}
The second vector in the dot product (\ref{eq:dot-product}) also has a bounded 2-norm. Using the definition of the norm and then Lemmas 3.1 and 3.2, we get
\begin{equation*}\label{eq:inequality-sequence-1}
    \begin{aligned}[b]
        \norm{\mu_{1:T}}^2 = \sum_{t=1}^{T} \frac{\widehat{m}_{t}^2}{t \,\widehat{v}_{t}} &\leq 1 + \sum_{t=2}^{T} \frac{1}{t} \left(K\,\frac{x_1^2}{x_2}\right)\\
        & \leq 1 + \left(K\,\frac{x_1^2}{x_2}\right)\int_{t=1}^{T}\,\frac{1}{t}\,dt\\
        & = 1 + K\,\frac{x_1^2}{x_2}\,\log T,
    \end{aligned}
\end{equation*}
and
\begin{equation}\label{eq:dot-2}
    \normtwo{\mu_{1:T}} \leq \sqrt{ 1 + K\,\frac{x_1^2}{x_2}\,\log T}.
\end{equation}

Therefore by (\ref{eq:dot-product}), (\ref{eq:dot-1}), (\ref{eq:dot-2}), and the Cauchy-Schwarz inequality, we have (\ref{eq:s-bound}).  
\end{proof}
The range of $\beta$ values that is permissible, given the conditions in Result 3.3, is shown in green in the next picture. For instance if $\beta_1 = 0.9$ then we must have $\beta_2 \in [0.99, 1)$. This range includes the $\beta$ values most commonly used in deep learning.\vspace{-0.2cm}
\begin{figure}[h!]
\centering
\includegraphics[scale = 0.5]{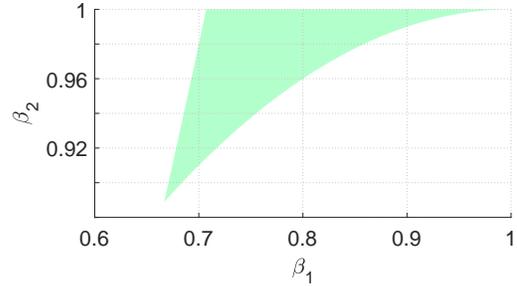}
\caption{Green shows the range of $\beta$ values that guarantees the upper bound on $s_T$ in (\ref{eq:s-bound}).}
\label{fig:beta-0-2}
\end{figure}
%\vspace{-0.2cm}

\section{Analysing Adam}

Our Result 3.3 generalizes an earlier result proven by Reddi and colleagues, namely Lemma 2 in \cite{reddi2019convergence}. Our methods of proof are quite different, but both approaches lead to bounds involving the quantity $\log T$ under a square-root sign. Reddi et al. proved their result for their AMSGrad optimizer, whereas ours holds for AMSGrad and for most or all common varieties of Adam itself, with or without bias correction (lines \ref{algo:mt-hat} and \ref{algo:vt-hat} of Algorithm~\ref{algo:adam-optimizer}) and with or without $\lambda$ variables (line \ref{algo:mt} of Algorithm~\ref{algo:adam-optimizer}) .

So Result 3.3 covers a wider range of optimizers than does Reddi and colleagues’ Lemma 2, but both their result and ours are very generally applicable in other ways. First, they both hold in a setting of online optimization that places no conditions except boundedness on the sequence of gradients, $g_{1:T}$. Second, they do not assume that the functions to be learned are convex, which is important if we aim to derive any conclusions about an optimizer’s performance in deep learning, where the loss landscape is usually far from convex, even in a small local neighbourhood of the network’s parameter vector. Third, both results concern processes that precede the parameter adjustments in the network, and consequently they make no assumptions about those adjustments. In particular, they do not require a mechanism that shrinks the learning rate factor as a function of time, as in line \ref{algo:thetat} of Algorithm~\ref{algo:adam-optimizer}. This point matters because even though most current analyses of Adam do require that the learning rate factor shrink with time, nonetheless in real deep-learning applications, performance is better without shrinkage.

All of this generality may be valuable, as we are unlikely to understand Adam or other Adam-type optimizers until we analyse them in settings other than convex optimization. Recent analyses in that convex setting have revealed a great many interesting properties of Adam-type optimizers, but nothing so far that explains these optimizers’ outstanding performance in deep learning. For example, one of the strongest results yet achieved is Reddi and colleagues' proof  \cite{reddi2019convergence} that AMSGrad zeroes average regret in the setting of online convex optimization. But its proven rate of convergence is not as good as that of simple gradient descent \cite{zinkevich2003online}, so this finding does not yet explain why AMSGrad works so much better than gradient descent in deep learning. For Adam, the case is even worse, as Reddi et al. \cite{reddi2019convergence} and Bock and Weiss \cite{bock2019non} have shown examples where Adam fails to zero the average regret. Strictly, Reddi and colleagues' example was of a failure when network parameters are optimized not by Adam alone but by Adam together with projection into a feasible set, but by adding weight decay it is straightforward to create a version of their example where pure Adam, without projection, also fails to zero the average regret.  

Another recent positive result is the proof by Bock and Weiss \cite{bock2021local} that Adam converges locally, meaning roughly that if it ever gets inside a convex neighbourhood of an optimum in parameter space then it will converge to that optimum. But again, the same is true of simple gradient descent, so this result shows only that Adam is as good as gradient descent in this respect, not that it is better.

Overall, then, the message seems to be that Adam is inferior to gradient descent in the setting of online convex optimization (OCO). The choice to analyse Adam in that setting goes back to Kingma and Ba \cite{kingma2014adam}, and it was reasonable because OCO shares with deep learning the crucial feature that the gradients change unpredictably from moment to moment.

But there are disanalogies, because in deep learning the gradients vary for two distinct reasons. First, they fluctuate from minibatch to minibatch. This effect, known as gradient noise, is probably well modelled by the random gradients of OCO. But second, the gradients of deep learning also drift as the network moves into new regions of parameter space where the local geometry of the loss function is different. So even without gradient noise — even with whole-batch as opposed to minibatch learning — the gradients would still vary unpredictably. This feature is not reflected in the OCO setting, and its absence may be preventing Adam-type optimizers from displaying their true worth. For instance, Reddi et al.\cite{reddi2019convergence} have shown that the reason Adam can fail in OCO is that its memory for past gradients is, in a certain sense, too short. But in deep learning, a short memory may let Adam discard gradient information that is obsolete because it belongs to regions of parameter space from which the network has already moved away.

\section{Conclusion}\label{sec:conclusion}

%the paper you wanted me to add: \cite{bock2021local} and \cite{bock2019non}
Our upper bound on $s_T$ in (\ref{eq:s-bound}) can replace the Kingma-Ba bound in analyses of the Adam optimizer.

%Our upper bound on $s_T$ in (\ref{eq:s-bound}) can replace the Kingma-Ba bound in analyses of the Adam optimizer. Our findings also resolve the clash between the attempted proof of convergence for Adam in \cite{bock2018improvement} and the failure of convergence demonstrated in \cite{reddi2019convergence}.

\section*{Appendix}\label{appendix}
\addcontentsline{toc}{section}{Appendices}
In our proof of Lemma 3.2, we said that $P(x_1) \geq 0$, where $P$ was the polynomial in (\ref{eq:time-derivative-2}), i.e.
\begin{equation}\label{eq:P(x1)}
    (1 - r)\,x_1^2 + r\,x_1 - 1 \geq 0,
\end{equation}
where $r \triangleq 2\log \beta_1 / \log \beta_2$. To verify (\ref{eq:P(x1)}), we observe that it is equivalent to
\begin{equation*}\label{eq:r-geq}
    r \geq \frac {1 - x_1^2}{x_1 - x_1^2},
\end{equation*}
which, by the definition of $r$, is in turn equivalent to
\begin{equation*}\label{eq:beta2-geq}
    \beta_2 \geq \beta_1^{\left(1 - \frac {\beta_1}{2 - \beta_1}\right)}.
\end{equation*}
Now given that $\beta_2 \geq 2\,\beta_1 - \beta_1^2$ in Lemma 3.2, it will suffice to show that
\begin{equation*}
    2\,\beta_1 - \beta_1^2 \geq \beta_1^{\left(1 - \frac {\beta_1}{2 - \beta_1}\right)}, 
\end{equation*}
i.e.
\begin{equation}\label{eq:y-leq}
    y(\beta_1) \triangleq \beta_1^{\frac {\beta_1}{\beta_1 - 2}} + \beta_1 - 2 \leq 0, 
\end{equation}
for $\beta_1 \in (0, 1)$.

Straightforward calculations show that
\begin{equation*}\label{eq:y(1)=0}
    y(1) = 0, ~~\frac {dy}{d\beta_1}(1) = 0, ~~\frac {d^2y}{d\beta_1^2}(1) = -2,
\end{equation*}
i.e. $y(1)$ is a strict local maximum.

To see that $y \leq 0$ in $(0, 1)$, we compute
\begin{equation}\label{eq:y-deriv}
     \frac {dy}{d\beta_1} = \beta_1^{\frac {\beta_1}{\beta_1 - 2}} \left(\frac {-2\log\beta_1 + \beta_1 - 2}{(2 - \beta_1)^2}\right) + 1 
\end{equation}
and observe that if $y$ were $0$ at any $\beta_1 \in (0, 1)$, then by (\ref{eq:y-leq}) the term
\begin{equation*}
     \beta_1^{\frac {\beta_1}{\beta_1 - 2}}
\end{equation*}
would $= 2 - \beta_1$, and (\ref{eq:y-deriv}) would become
\begin{equation*}
     \frac {dy}{d\beta_1} = \frac {-2\log\beta_1}{2 - \beta_1} > 0. 
\end{equation*}
So if $y$ were $\geq 0$ at any $\beta_1^\prime \in (0, 1)$ then it would stay $\geq 0$ on $(\beta_1^\prime, 1)$, contradicting the fact that $y(1)$ is a strict local maximum.  Therefore $y$ must remain $\leq 0$ in $(0, 1)$, confirming (\ref{eq:y-leq}) and (\ref{eq:P(x1)}).   

\bibliographystyle{unsrt}
\bibliography{reffs.bib}

\end{document}